\documentclass[letterpaper,english]{article}
\PassOptionsToPackage{natbib=true}{biblatex}
\PassOptionsToPackage{hyphens}{url}
\usepackage[T1]{fontenc}
\usepackage[latin9]{inputenc}
\usepackage{babel}
\usepackage{array}
\usepackage{refstyle}
\usepackage{booktabs}
\usepackage{url}
\usepackage{multirow}
\usepackage{graphicx}
\usepackage{rotating}
\usepackage[unicode=true,pdfusetitle,
 bookmarks=true,bookmarksnumbered=false,bookmarksopen=false,
 breaklinks=false,pdfborder={0 0 1},backref=false,colorlinks=false]
 {hyperref}

\makeatletter

\AtBeginDocument{\providecommand\Figref[1]{\ref{Fig:#1}}}
\AtBeginDocument{\providecommand\Tabref[1]{\ref{Tab:#1}}}
\pdfpageheight\paperheight
\pdfpagewidth\paperwidth

\providecommand{\tabularnewline}{\\}
\RS@ifundefined{subsecref}
  {\newref{subsec}{name = \RSsectxt}}
  {}
\RS@ifundefined{thmref}
  {\def\RSthmtxt{theorem~}\newref{thm}{name = \RSthmtxt}}
  {}
\RS@ifundefined{lemref}
  {\def\RSlemtxt{lemma~}\newref{lem}{name = \RSlemtxt}}
  {}

\newcommand{\lyxaddress}[1]{
	\par {\raggedright #1
	\vspace{1.4em}
	\noindent\par}
}

\@ifundefined{showcaptionsetup}{}{%
 \PassOptionsToPackage{caption=false}{subfig}}
\usepackage{subfig}
\makeatother

\usepackage[style=authoryear]{biblatex}
\begin{document}
\title{Provident Vehicle Detection at Night: \\
The PVDN Dataset}
\author{Lars Ohnemus,$^{1,}$\thanks{The research was performed during employment at Dr.\ Ing.\ h.c.\ F.\ Porsche
AG.}~$^{,}$\thanks{Authors contributed equally.} ~Lukas Ewecker,$^{2,\dagger}$
Ebubekir Asan,$^{2,\dagger}$\\
Stefan Roos,$^{2}$ Simon Isele,$^{2}$ Jakob Ketterer,$^{1,*}$\\
Leopold Müller,$^{1,*}$ and Sascha Saralajew$^{3,4,*,\dagger}$}
\date{\vspace{-5ex}
}
\maketitle

\lyxaddress{\begin{center}
$^{1}$Karlsruhe Institute of Technology, Karlsruhe, Germany\\
$^{2}$Dr.\ Ing.\ h.c.\ F.\ Porsche AG, Weissach, Germany\\
$^{3}$Leibniz University Hannover, Institute of Product Development,
\\
Hannover, Germany\texttt{}~\\
$^{4}$Bosch Center for Artificial Intelligence, Renningen, Germany\texttt{}~\\
\texttt{}~\\
\texttt{\{ohnemus.lars, sascha.saralajew\}@gmail.com}
\par\end{center}}

\begin{abstract}
For advanced driver assistance systems, it is crucial to have information
about oncoming vehicles as early as possible. At night, this task
is especially difficult due to poor lighting conditions. For that,
during nighttime, every vehicle uses headlamps to improve sight and
therefore ensure safe driving. As humans, we intuitively assume oncoming
vehicles before the vehicles are actually physically visible by detecting
light reflections caused by their headlamps. In this paper, we present
a novel dataset containing 59\,746 annotated grayscale images out
of 346 different scenes in a rural environment at night. In these
images, all oncoming vehicles, their corresponding light objects (e.\,g.,
headlamps), and their respective light reflections (e.\,g., light
reflections on guardrails) are labeled. This is accompanied by an
in-depth analysis of the dataset characteristics. With that, we are
providing the first open-source dataset with comprehensive ground
truth data to enable research into new methods of detecting oncoming
vehicles based on the light reflections they cause, long before they
are directly visible. We consider this as an essential step to further
close the performance gap between current advanced driver assistance
systems and human behavior.
\end{abstract}

\section{Introduction}

At night, provident perception of oncoming vehicles is essential for
safe driving. The risk of fatal accidents is reduced by detecting
oncoming vehicles early. Humans are capable of provident vehicle detection
at night through predicting the position of other road users through
their headlamps' light glares and reflections long before the vehicles
are directly visible. However, those light-artifacts are challenging
to separate from other light sources within a typical rural scenery,
so current vehicle detection systems only identify directly visible
headlamps and do not take this natural human behavior into account.
Besides diver assistance cameras, other sensor types commonly used
for Advanced Driver Assistance Systems (ADAS), like LIDAR,\footnote{Light detection and ranging.}
are not even capable of detecting those features because those do
not correlate to a physical object. Therefore, camera systems might
be the only possible sensors appropriate to providently detect vehicles
at night. 

The provident detection of oncoming vehicles at night could be a key
feature for new ADAS functionalities---for example, to predictively
control the car's high beam system before the oncoming vehicle is
in direct sight. In addition to this obvious use-case, many other
use-cases for such a provident detection system are possible. For
example, at night, autonomous driving systems could use the information
to reduce speed before tight bends or to notice other vehicles at
intersections without direct visibility. Therefore, early detection
could remove many causes of car accidents at night.

It may come to mind to use machine learning techniques to build a
capable perception system for provident vehicle detection at night
based on camera images. For this, convolutional neural networks and
similar architectures have proven their superiority against other
pattern recognition methodologies \citep[e.\,g.,][]{fasterrcnn,resnet,yolo3}.
However, this superior performance comes at a price: A representative
dataset for the task has to be recorded and evaluated. Any discrepancy
between the acquired dataset and the application domain will be noticeable
in the system's real-world performance. This hurdle may not be too
challenging for most tasks, but it is apparent for the described problem.
At this point, no dataset for provident vehicle detection at night
is publicly available. Even the representation and annotation method
for light-features is up to debate. The task of predicting oncoming
vehicles can be represented in various ways. For example, it is possible
to assign each camera frame a label to describe whether a car is approaching.
Phrasing the problem as a classification task does allow for a simple
annotation but only a pure end-to-end approach, which is usually not
suitable for ADAS since the system behavior would be not comprehensible.
A more sophisticated way would be to apply commonly used object detection
methods. In earlier work, we showed that bounding boxes around light-features
can be used to train machine learning models \citep{Oldenziel2020}.
But still, the problem of the end-to-end approach remains. Also, the
description via bounding boxes is a non-intuitive restriction to the
fuzzy nature of such light-artifacts.

\subsubsection{Contributions}

This work analyzes the problem of ``provident detection of vehicles
at night'' and gives insights into the development process of a suitable
dataset. The derived dataset is called ``PVDN'' (Provident Vehicle
Detection at Night) and is designed in a way that allows the usage
of various representation strategies. Three different representations
are presented and discussed. To accelerate development, the PVDN dataset
is made publicly available.\footnote{\url{https://www.kaggle.com/saralajew/provident-vehicle-detection-at-night-pvdn}}
Furthermore, source code is released to provide an easy interface
to the dataset in Python and PyTorch.\footnote{\url{https://github.com/larsOhne/pvdn}}

\subsubsection{Outline}

The paper is structured as followed: First, a summary of related work
is given. Building on the knowledge from  previous work, a more general
annotation strategy is presented. The strategy is then applied to
a large number of preselected video sequences. Here, the annotation
process is discussed, and the resulting PVDN dataset is analyzed.
To conclude, the results as well as potential future concepts are
discussed.

\section{Related work}

Besides the fact that almost every car manufacturer offers a camera-based
perception system to perceive oncoming vehicles at night---for example,
to control the car's high beams---the problem to providently detect
oncoming vehicles based on their emitted light is not studied. This
might be caused by the state-of-the-art vehicle detection systems
at night that are based on blob detection systems followed by blob
pairing and classification \citep[e.\,g.,][]{Juric.2014,Lopez.2008,Sevekar.2016,P.F.Alcantarilla.2011,Eum.2013}.
Of course, these systems are impressively efficient in terms of computational
requirements to detect headlamp blobs at night but also induce a significant
restriction: The detection of blobs. Obviously, light reflections
in the environment caused by oncoming vehicles cannot be sufficiently
described by blobs.

In previous work \citep{Oldenziel2020}, we performed a test group
study to specify the discrepancies between current ADAS and human
provident behavior. Additionally, as a proof of concept, we showed
that the provident detection of vehicles at night is possible when
using modern machine learning techniques like an adapted Faster-RCNN
\citep{fasterrcnn} architecture.

\section{The PVDN dataset}

The PVDN dataset is derived from a test group study that investigated
humans' provident vehicle detection capabilities and consists of \emph{annotated
grayscale camera images}. During the study, the onboard camera of
the test car was used to capture grayscale images of two different
exposure cycles. The resulting image cycles are called ``day cycle''
(short exposure) and ``night cycle'' (long exposure). Image sequences
of scenes (available through the test group study) are selected according
to the appearance of oncoming vehicles and padded to create parts
without vehicles in them. Mostly, such sequences begin when there
is no sign of an oncoming vehicle yet (no annotation) and end when
the vehicle has passed.

The main objective of this dataset is to divide the complex task of
predicting vehicle positions at night without direct sight into single
subtasks. Similar to human behavior, these subtasks consist of perceiving
light-artifacts and predicting the position or occurrence of an oncoming
vehicle. Therefore, we aim to design a dataset that provides rich
information about features that we humans would use to detect oncoming
vehicles at night. In the following, we describe in more detail how
the dataset was derived and how the images of the scenes are annotated.

\subsection{Terminology}

A few definitions for clarification:
\begin{description}
\item [{Scene}] A scene is a sequence of image frames. In each sequence,
at least one oncoming vehicle can be found. All scenes were created
from captured video frames obtained during the test group study.
\item [{Direct\,/\,indirect}] A vehicle is considered directly visible,
when its headlamps are visible. Vehicles where only glares or reflections
indicate its position are considered indirect.
\item [{Instance}] All spatially bound light artifacts (e.\,g., reflections,
glares, or direct headlamps) are called ``instance'' of a certain
vehicle.
\end{description}

\subsection{Hierarchy}

\begin{figure}
\begin{centering}
\includegraphics[width=0.8\columnwidth]{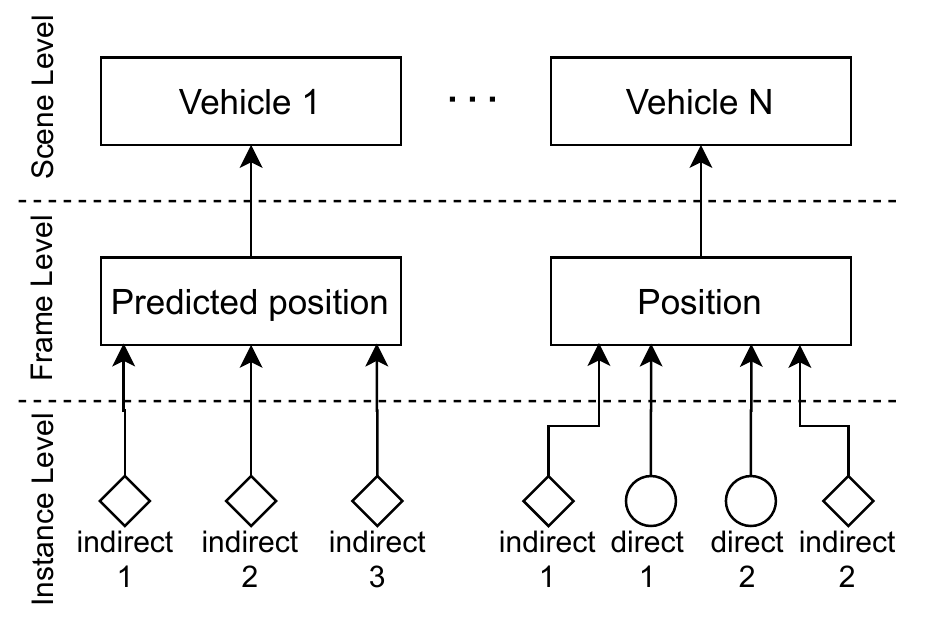}
\par\end{centering}
\caption{\label{fig:Hierachy-of-different}Hierarchy of different entities
used for the PVDN dataset.}
\end{figure}
The information we want to retrieve from the raw image data spans
multiple hierarchical levels. On the highest semantic level, the aim
is to predict oncoming vehicles. To maintain generality, we assume
that each scene may contain multiple vehicles. For each frame, the
position of the vehicles needs to be specified. At this frame-level,
the vehicles are detected without the time-coherence derived from
a complete scene inspection. Since the vehicle position itself is
an abstract notion for not directly visible vehicles, it can sometimes
only be perceived through light artifacts within a frame. Those instances
build up the knowledge about the vehicle and are therefore placed
on the lowest semantic level. \Figref{Hierachy-of-different} shows
the relationship between those different entities. It should be noted
that valuable information could be lost if only the vehicle position
is annotated.

The proposed hierarchy yields two requirements for a suitable dataset: 
\begin{enumerate}
\item Both the vehicle position and the related instances need to be annotated,
and
\item the annotations should be coherent over multiple frames to capture
the temporal characteristics of an oncoming vehicle.
\end{enumerate}

\subsection{Vehicle Position}

\begin{figure}
\begin{centering}
\subfloat[\label{fig:Direct-Vehicle-Position}Direct vehicle position]{\begin{centering}
\includegraphics[width=0.8\columnwidth]{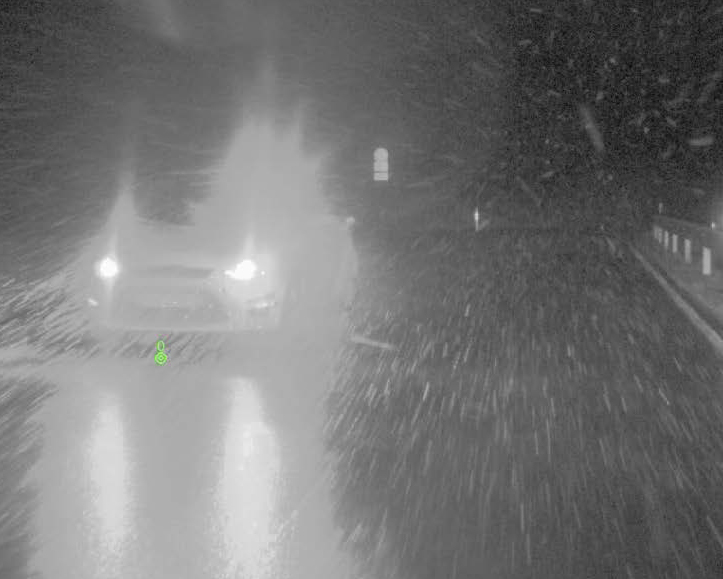}
\par\end{centering}
}
\par\end{centering}
\begin{centering}
\subfloat[\label{fig:Indirect-Annotations}Indirect annotations]{\begin{centering}
\includegraphics[width=0.8\columnwidth]{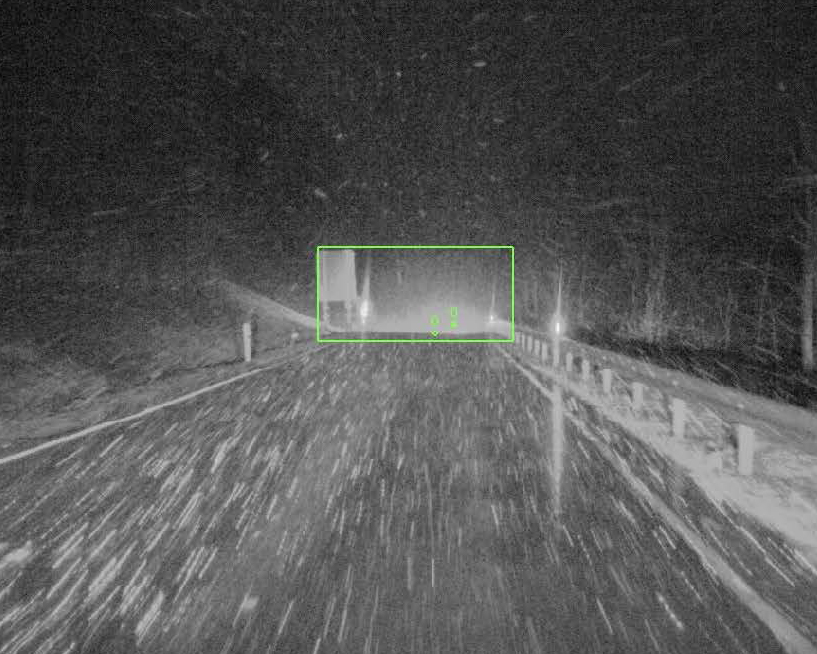}
\par\end{centering}
}
\par\end{centering}
\caption{Annotation examples: \ref{fig:Direct-Vehicle-Position} shows an annotation
of a direct vehicle position; \ref{fig:Indirect-Annotations} shows
an indirect instance (glare) and a indirect vehicle position as well
as an automatically inferred bounding box.}
\end{figure}

The annotation of an approaching vehicle poses two challenges:
\begin{enumerate}
\item How to represent the actual position within a frame, and
\item where to place the vehicle position when it is not visible yet. 
\end{enumerate}
In the first case, the most efficient way to annotate the position
is to place a keypoint at a remarkable point on the vehicle. We used
a keypoint placed on the road centrally between both headlights (e.\,g.,
\Figref{Direct-Vehicle-Position}). The actual size of the vehicle
would be redundant information since it can always be inferred from
the vehicle position and the two headlight instances. Stringently,
in the second case, the position of indirect vehicle positions is
annotated with a keypoint as well (see \Figref{Indirect-Annotations}).
So if the vehicle itself is not visible, the keypoint is placed at
that point on the visible road, at which the vehicle will appear first.
If the road touches the edge of the image, the intersection point
is used. This approach has some key advantages: 
\begin{itemize}
\item It conserves temporal coherence when transitioning between indirect
and direct vehicle positions, and
\item the keypoint can be directly used as a target when predicting the
occurrence position of vehicles.
\end{itemize}
Additionally, a Boolean label is added to each vehicle position to
specify whether it is direct or indirect.

\subsection{Instances}

To retrieve the maximal information content within a frame, light-artifacts
are annotated as well. In addition to the challenges for vehicle positions,
the fuzzy nature of the light-artifacts needs further thought. While
direct instances are well specified (headlights), the total number
and size of the indirect instances cannot be specified objectively.
Therefore, we introduce the concept of salient keypoints. For each
spatially bound light-artifact, the intensity maximum is annotated
with a keypoint. This correlates nicely with the human perception,
looking around in the scene to find salient areas. So the keypoints
put onto indirect instances can be viewed as carefully selected eye-fixation
points. While this cannot be seen as absolute ground truth, it still
gives hints towards the attention of human annotators. These keypoints
can be used to derive more general and objective representations than,
for example, hand-drawn bounding boxes. Also, for direct instances
this annotation strategy translates nicely since the keypoint can
be placed centrally within the headlight cone. Additionally, a Boolean
label is added to each instance keypoint to specify whether it is
direct or indirect.

\subsection{Annotations methodology}

As already mentioned, the annotation of indirect instances and vehicle
positions is subjective. To minimize the subjectiveness effects, an
iterative procedure for the annotation process was chosen. Keypoints
for vehicle position and instances were annotated with a custom annotation
tool. Then, the annotations were reviewed by multiple annotators and
the placement and number of keypoints was discussed. Additionally,
a set of guidelines for different conditions and circumstances was
given to the annotators. The reviewed annotations were then corrected
by an annotator. An additional comment feature within the annotation
tool ensures ease of communication. Simple replacement patterns are
used as well to unify different annotation policies. To balance the
different camera exposures, approximately for half of the scenes the
night cycle was annotated. The custom annotation tool is also publicly
available on GitHub.\footnote{\url{https://github.com/larsOhne/pvdn}}

\subsection{Statistics}

\begin{table}
\begin{centering}
\begin{tabular}{cccccc}
\toprule 
 &  & \begin{turn}{90}
\# of scenes
\end{turn} & \begin{turn}{90}
\# of images
\end{turn} & \begin{turn}{90}
\# vehicle positions
\end{turn} & \begin{turn}{90}
\# instances
\end{turn}\tabularnewline
\midrule
\midrule 
\multirow{2}{*}{train} & day & 113 & 19\,078 & 15\,403 & 45\,765\tabularnewline
\cmidrule{2-6} \cmidrule{3-6} \cmidrule{4-6} \cmidrule{5-6} \cmidrule{6-6} 
 & night & 145 & 25\,264 & 26\,615 & 72\,304\tabularnewline
\midrule 
\multirow{2}{*}{validation} & day & 20 & 3\,898 & 2\,602 & 7\,244\tabularnewline
\cmidrule{2-6} \cmidrule{3-6} \cmidrule{4-6} \cmidrule{5-6} \cmidrule{6-6} 
 & night & 25 & 4\,322 & 3\,600 & 12\,746\tabularnewline
\midrule 
\multirow{2}{*}{test} & day & 19 & 3\,132 & 3\,045 & 9\,338\tabularnewline
\cmidrule{2-6} \cmidrule{3-6} \cmidrule{4-6} \cmidrule{5-6} \cmidrule{6-6} 
 & night & 24 & 4\,052 & 3\,384 & 10\,438\tabularnewline
\midrule 
\multirow{3}{*}{total} & day & 152 & 26\,108 & 21\,050 & 62\,347\tabularnewline
\cmidrule{2-6} \cmidrule{3-6} \cmidrule{4-6} \cmidrule{5-6} \cmidrule{6-6} 
 & night & 194 & 33\,638 & 33\,599 & 95\,488\tabularnewline
\cmidrule{2-6} \cmidrule{3-6} \cmidrule{4-6} \cmidrule{5-6} \cmidrule{6-6} 
 & cumulated & 346 & 59\,746 & 54\,649 & 157\,835\tabularnewline
\bottomrule
\end{tabular}
\par\end{centering}
\caption{\label{tab:Dataset-split}Dataset split.}

\end{table}
The dataset contains 59\,746 annotated images, spread over 346 scenes.
In comparison to the Enz-dataset \citep{Oldenziel2020}, the share
of indirect instances to direct instances is better balanced (ca.\ 43\%
of instances are indirect). The same is true when considering the
detections of indirect vehicles (51\% of the annotated vehicle positions
are indirect). The total numbers of images, scenes and instances are
summarized in \Tabref{Dataset-split}.

\section{Conclusion}

In this paper, we have introduced the PVDN dataset that is conducted
to evaluate and train algorithms to providently detect vehicles at
night. Based on this annotated dataset---in accordance with the publication---we
hope to provide a baseline to study ``object'' detection frameworks
that are closer to human capabilities at night. Additionally, as discussed
before, the proposed task gives good reasons why keypoint prediction
frameworks are better suited in this case. However, the current trend
in machine learning focuses more on bounding box predictions that
might be cumbersome for light-artifacts. Finally, as most object detection
benchmarks datasets focus on daylight scenarios, this dataset is a
good starting point to leverage object detection frameworks at night
to the next level---as it is long overdue.

\printbibliography

\end{document}